\documentclass[runningheads]{llncs}
\pdfoutput=1
\usepackage{graphicx}

\usepackage{tikz}
\usepackage{comment}
\usepackage{amsmath,amssymb} 
\usepackage{color}

\usepackage{pgfplots}
\usepackage{float}
\usepackage{booktabs,amsfonts,dcolumn}
\usepackage{cuted}
\newcolumntype{d}[1]{D..{#1}}

\begin{document}

\title{A General-Purpose Dehazing Algorithm based on Local Contrast Enhancement Approaches} 

\titlerunning{A General Purpose Dehazing Algorithm based on LCE Approaches}
%
\author{Bangyong Sun\inst{1,2}
  \and
  Vincent Whannou de Dravo\inst{1}
  \and
  Zhe Yu\inst{1}
}
\authorrunning{B. Sun et al.}
%
\institute{School of Printing, Packaging and Digital Media, Xi’an University of Technology, Xi\'an, China\\
  \email{\{sunbangyong\}@xaut.edu.cn}
\and
Key Laboratory of Spectral Imaging Technology CAS, Xi’an Institute of Optics and Precision Mechanics, Chinese Academy of Sciences, Xi’an, China\\
}
\maketitle
\begin{figure}[ht]
  \begin{center}
  \centering
  \includegraphics[width=.99\textwidth]{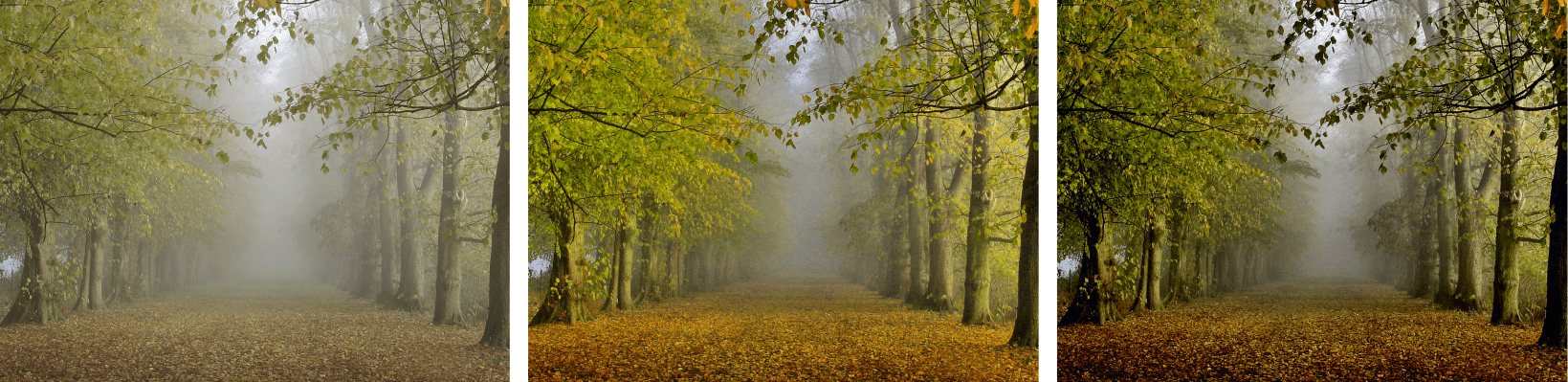}
   \caption{\textbf{An Example of our novel and fast Dehazing Algorithm:} Forest image on hazy condition (left), and output images of our method with two different settings (middle \& right). The one in the middle works with homogeneous haze mostly($\lambda$ is fixed), and the other works with homogeneous and non-homogeneous haze. We use clahe \cite{Zuiderveld:1994} for these two outputs.}
\end{center}%
\end{figure}

\begin{abstract}
Dehazing is in the image processing and computer vision communities, the task of enhancing the image taken in foggy conditions. To better understand this type of algorithm, we present in this document a dehazing method which is suitable for several local contrast adjustment algorithms. We base it on two filters. The first filter is built with a step of normalization with some other statistical  tricks while the last represents the local contrast improvement algorithm. Thus, it can work on  both CPU and GPU for real-time applications. We hope that our approach will open the door to new ideas in the community. Other advantages of our method are first that it does not need to be trained, then it does not need additional optimization processing. Furthermore, it can be used as a pre-treatment or post-processing step in many vision tasks. In addition,  that it does not need to convert the problem into a physical interpretation, and finally that it is very fast. This family of defogging algorithms is fairly simple, but it shows promising results compared to state-of-the-art algorithms based not only on a visual assessment but also on objective criteria.
\dots
\end{abstract}

\section{Introduction}
Haze, fog, dust, mist or smoke represent a major visibility issue in outdoor or indoor computer vision applications. In fact, these natural or artificial phenomena degrade the quality of images and it is not possible to process images without providing a solution to these issues. Thus, solving  dehazing problems has attracted many researchers in the field of computer vision in the past two decades.

Dehazing algorithms  can be roughly divided into two categories: learning approaches and non-learning approaches. Earlier work focuses on non-learning based approaches,  while later work more likely addresses learning approaches. Furthermore, we can even divide non-learning techniques in two categories: single image dehazing v. multiple inputs dehazing. Multiple inputs dehazing techniques appear in the earliest \cite{Nayar:1999}, \cite{Narasimhan:2000}, \cite{Narasimhan:2002}, \cite{Narasimhan:2003}, but they may also appear in more recent papers \cite{Caraffa:2012}. Single image dehazing pioneering work come with Tan \cite{Tan:2008} and Fattal \cite{Fattal:2008}. While Tan based his work on a contrast enhancement technique, Fattal used a method based on the haze physical model introduced by Koschmieder. The following single image dehazing techniques are either based on a contrast enhancement method \cite{Tarel:2009}, \cite{Ancuti:2013}, \cite{Meng:2013}, \cite{Choi:2015} or on physical techniques\cite{He:2011}, \cite{Kratz:2009}, \cite{Tang:2014}, \cite{Fattal:2014}, \cite{Berman:2016}, \cite{cho:2018}. Learning approaches essentially used supervised machine learning techniques. Some notable algorithms are in \cite{Cai:2016}, \cite{Li:2017}, \cite{Ren:2018}, \cite{Li:2018}, \cite{Wang:2018}.

While these algorithms are effective, the mechanism that makes them efficient is often unclear,  and they could not be unified in a single framework. In this paper, we provide a general framework to enhance the images taken in hazy or foggy conditions. This family of algorithms consists of two filters. The first filter employs a simple normalization with a statistical artifice on the input hazy image, and it tends to darken and homogenize the haze density over the entire image. In addition, it is implemented in a per-pixel fashion. While the second filter goal is to remove the remaining haze from the output of the first filter stage based on an image local contrast enhancement scheme.
The fundamental aims of our novel algorithm are first to propose a family of algorithms for dehazing based on well-known image enhancement algorithms. Second, the possibilities for this family of algorithms to work on video processing tasks for those that have a fast subroutine. Third, because of its speed both on CPUs and GPUs, it can be used as a step of pre or post-treatment in many vision applications. Other gains are i) It does not need any training to function, ii) It does not need an optimization trick to give amazing results iii) The principle behind its implementation is quite straightforward iv) It is not a physical based method.
We are the first or among the first presenting such a general, simple and effective framework using a set of local contrast enhancement (LCE) algorithms.
Our technique based on LCE routines, resemble to the work of \cite{Tan:2008}, \cite{Tarel:2009}, \cite{Ancuti:2013}, \cite{Meng:2013} or \cite{Choi:2015}, \cite{Galdran:2015}, \cite{Galdran:2015b}, \cite{Galdran:2017}. However, in contrast to some of these state-of-the-art methods, ours appears  simple and more general-purpose. Also, our algorithm clearly outperform these state-of-the-art algorithms in many hazy circumstances. Our approach is different from the work of \cite{VazquezCorral:2018}, because their work is based on global constrained histogram flattening scheme, whereas ours is tested on many LCE algorithm, and its pipeline is different also. It is worth noting that the algorithm in \cite{Galdran:2017} has also a general-purpose behaviour, and that ours does not work with pure Retinex algorithms \cite{Land:1971}, \cite{Jobson:1997}, \cite{Provenzi:2007}, but does work with Retinex-like algorithms that incorporate a spatial high-pass filter component.

\section{Related Work}
Visibility is a crucial indicator for understanding and dealing with the interior and exterior scene in vision applications.
Image defogging algorithms can be classified into many categories depending on the type of input that provides the final result.

Earliest work on the problem of scene recovery in bad weather condition are based on estimating the scene depth in order to recover the scene most useful properties. The work of \cite{Nayar:1999}, \cite{Narasimhan:2002}  are notable advances in this categories. 
 In \cite{Nayar:1999}, the algorithm exploits the direct transmission to estimate the relative depths of the light sources from two images taken under different scattering coefficients at night. The work in \cite{Narasimhan:2002} set the goal to compute absolute depth  map from image taken in scattering media. 
Another type of algorithm principally works on multiple input images in order to solve the dehazing task \cite{Narasimhan:2000}, \cite{Narasimhan:2003}, \cite{Caraffa:2012}, \cite{Li:2015}.
For instance, Narasimhan and Nayar \cite{Narasimhan:2000} introduce a method which uses two images under different weather conditions to derive the haze-free image and the depth information to reconstruct the 3D scene. To achieve this, they assume that both images share the same color of atmospheric light, but different color of direct transmission is speculated. Later on, the same authors in \cite{Narasimhan:2003} develop another algorithm based on multiple images.
A third type of algorithm works on polarizing filters \cite{Schechner:2001}. 
In \cite{Schechner:2001}, the authors established a relationship between the image formation process and the polarization effects of atmospheric veil. The problem is then solved with two or more images through a polarizer at different orientations.  
Known depth based dehazing algorithm is another singular approach \cite{Oakley:1998}, \cite{Narasimhan:2003b}, \cite{Hautiere:2007}, \cite{Kopf:2008}. In this category, Oakley and Satherley \cite{Oakley:1998} design a single defogging algorithm into two steps. In the first, they estimated parameters from a physical model. In the final step, they used the parameters for contrast enhancement.
Single image haze removal capital work were the work of Tan \cite{Tan:2008} and the work of Fattal \cite{Fattal:2008}. Tan solved the defogging problem thanks to two assumptions. First he assumed that the atmospheric light has the brightest value. Second, he hypothesized that the scene reflection has the maximum contrast. On the other hand, Fattal  fixed the problem by presuming that the shading and the transmission are locally and statistically uncorrelated. 
He \textit{et al.} \cite{He:2011} proposed a distinguished prior called the dark channel. The dark channel is based on a key observation - most local patches in haze-free outdoor images contain some pixels which have very low intensities in at least one color channel. Using this new assumption, they are able to estimate  the thickness of the haze and recover a high quality haze-free image.
Tarel and Hauti{\`e}re \cite{Tarel:2009}, as Tan Did, also made the maximal contrast assumption. They then hypothesized that the airlight is normalized and upper bounded.
Ancuti and Ancuti \cite{Ancuti:2013} bring forward a new dehazing algorithm based on the Gray-world colour constancy and a global contrast enhancement method. Their method did not rely on a physical meaning of the problem.
Meng \textit{et al.} \cite{Meng:2013} enlarged the concept of dark channel prior by applying it to the transmission map. Tang \textit{et al.} \cite{Tang:2014} used a machine learning crude features to find out a solution of the transmission map. 
Fattal \cite{Fattal:2014} suggested a new prior, the color lines, and he derived a local formation model that explains the color-lines in the context of hazy scenes. In contrary to other previous work, the assumption is not applied to the entire image, thus this  trick allowed the approach to have more success than previous methods.
Choi \textit{et al.} \cite{Choi:2015} have introduced DEFADE, a perceptual image defogging method based on image Natural Scene Statistics (NSS) and fog aware statistical features. DEFADE achieves better results on darker, denser foggy images as well as on standard defog test images than state-of-the-art dehazing algorithms.
Berman and colleagues  \cite{Berman:2016}, as Fattal did, contemplated the solution over an adaptive prior that they called haze-line. In that case as well, their algorithm was able to avoid artifacts found in many state-of-the-art approaches.
Galdran \textit{et al.} \cite{Galdran:2017} provided a rigorous mathematical proof of the dual relation linking the problems of image dehazing and non-uniform lighting separation, showing that the application of a Retinex operation on an inverted image followed by an inversion of the result provides again a dehazed result, and vice versa.
Cho \textit{et al.} \cite{cho:2018} mixed a model-based method and a fusion-based method, where the input images were broken down into intensity and Laplacian modules for pixel and level enhancement degraded to improve images taken in hazy conditions. Their concept was a further test for single image comparison and for robots vision.\\
A sub-category of single image dehazing algorithm are algorithms that use machine learning supervised model to learn parameters of dehazing Equation or generate the haze-free image directly. Among this category, we can mention the work of \cite{Cai:2016}, \cite{Li:2017}, \cite{Ren:2018}, \cite{Li:2018}, \cite{Wang:2018}.
Cai \textit{et al.} \cite{Cai:2016} presented a trainable end to end deep learning network, DehazeNet. The network was designed to learn  the medium transmission map that it is latterly use to recover a new hazy image.
In \cite{Li:2017}, Li \textit{et al.} introduced a new end-to-end  design based on re-formulated atmospheric scaterring model called All-in-One Dehazing Network (AOD-Net). 
Ren \textit{et al.} \cite{Ren:2018} trained their auto-encoder network with three different transform plus the original hazy image: white balance, contrast enhancement and  Gamma correction. 
Li \textit{et al.} \cite{Li:2018} introduced a new haze removal algorithm based on conditional generative adversarial networks (cGANs) approach.
After having demonstrated the importance of luminance channel in the YCbCr color space, Wang and colleagues \cite{Wang:2018}introduced the Atmospheric illumination Prior over a CNN network called AIPNet for short.

\section{Proposed Approach}
Dehazing problems can be solved with various techniques. One common way of doing this is to use Koschmieder physical model:
\begin{equation}
\label{eq:01}
I(\bold{x})=J(\bold{x})t(\bold{x}) + A(1-t(\bold{x}))
\end{equation}

Where $\bold{x}$ is a pixel (single-pixel or not). The hazy image is the sum of the scene's radiance $J(\bold{x})$ and the atmospheric light $A$, weighted by a transmission factor $t(\bold{x})$. The $A$ is the airlight scattered by an object located at infinity with respect to the observer.
Here, we use another way to solve the problem. Our method relies on local contrast improvement techniques, it does not require any post-processing or any optimization procedure to function.

Our dehazing algorithm has been tested on some local contrast enhancement scheme, and it clearly shows satisfying results. Thus, it represents a family of dehazing algorithms that deals with  local contrast filtering algorithms such as \cite{Harris:1977}, \cite{Narendra:1981}  ace \cite{Gatta:2002}, \cite{Rizzi:2003}, stress \cite{Kolas:2011} or clahe \cite{Zuiderveld:1994}. It is a very fast algorithm when the local contrast subroutine is also fast. It can enhance both dense and non-dense images taken in hazy or foggy conditions. However, it appears to be more suitable for non-dense conditions. This simple and novel algorithm is composed of two steps of filtering. The front-end filter that uses a simple normalization step combined with a statistical trick, and it tends to darken and homogenize the haze density over the entire image. Let $\mathbf{x}_i$ denote a given sample in the image, $\mathbf{x}_{min}$ and $\mathbf{x}_{max}$ are respectively the minimum and the maximum sample  
in the entire image. The initial filter is computed in two stages as follows:
\begin{eqnarray}
  \label{InitEq}
  f_{init} (\mathbf{x}_i) =  \frac{\mathbf{x}_i - \mathbf{x}_{min}}{\mathbf{x}_{max} - \mathbf{x}_{min}}
\end{eqnarray}

Before applying the second filter, we further extend the previous filter to the more general filters as the following:
\begin{eqnarray}
  \label{general_filter}
  f_g (\mathbf{x}_i) = f_{init} (\mathbf{x}_i) - \lambda(\mathbf{x}_i)\lVert f_{init}(\mathbf{x}_i) \rVert
\end{eqnarray}

Where the function $\lambda$ is either a constant $\in$ $\/[0, 1\/]$ or any other specified function. In our settings, we have tested $\lambda = 0.35$ and the inverted intensity function of the first filter, that is:
\begin{eqnarray}
  \label{invert_func}
  \lambda(\mathbf{x}_i) = 1 - f_{init} (\mathbf{x}_i)
\end{eqnarray}
One can notice here that the inverted intensity function is also $\in$ $\/[0, 1\/]$. $\lVert . \rVert$ represents the  distance of a given pixel over the three chromatic channels. We have only investigated Euclidean norm in this paper, but the other norm may be interesting also.

We experimentally notice that when Equation \ref{invert_func} is used in Equation \ref{general_filter}, then it can serve as a dehazing algorithm directly. Because its appearance might look too dark for some images, the contrast enhancement techniques or a gamma correction algorithm may be necessary to have the final output.

Empirically, we notice that the general form of Equation \ref{InitEq} is as the following:
\begin{eqnarray} 
  f_{init} (\mathbf{x}_i) =  \alpha \frac{\mathbf{x}_i - \mathbf{x}_{min}}{\mathbf{x}_{max} - \mathbf{x}_{min}} + \beta
\end{eqnarray}
$\alpha$ and $\beta$ are two real numbers. In the experiment presented here, we have set $\alpha=1.0$ and $\beta=0.0$.

The back-end and final filter  is an image local contrast enhancement scheme that takes the output of the front-end filter and it can be expressed as follows:
\begin{eqnarray} 
  f_{final} (\mathbf{x}_i) = f_{LCE}(f_g (\mathbf{x}_i) )
\end{eqnarray}
 Where $f_{LCE}$ represents a local contrast enhancement algorithm such as clahe. We have empirically checked that $f_{LCE}$ functions share a local contrast property (see supplementary documents for more details on this).
 One can notice that the initial and the final filter algorithms work on global and local contrast schemes respectively, and the algorithm does not require an explicit segmentation as well. This family of algorithm can appear to be useful even in some challenging cases such as dense haze removal problem. We have use two main settings in our research. The first one ($\lambda$ is fixed) works with homogeneous haze. The second ($\lambda$ is dynamic) not  only fits homogeneous haze, but meets also  non-homogeneous haze.

This stage dehazing can be further augmented with a haze physical constraint procedure to tackle white balancing issue by using soft matting \cite{Levin:2006}, \cite{Hsu:2008} to refine the transmission in the model. We refer the reader to the supplementary materials for more on this.

\section{Experiment and Results}
We did our test on Linux platforms. We perform an assessment both subjectively and objectively. For this experiment, we are considering images from different databases of state-of-the-art approaches. We first visually compare results of the outputs of our family of algorithms with each other. Then, a visual comparison is  made between the family of algorithms with some state-of-the-art approaches. Finally, we use two objective metrics \cite{Choi:2015}, \cite{Hautiere:2008} to compare the outputs of our set of algorithms with advanced literature techniques.
\begin{figure*}[h!]
  \begin{center}
        \centering 
                \includegraphics[width=0.99\textwidth]{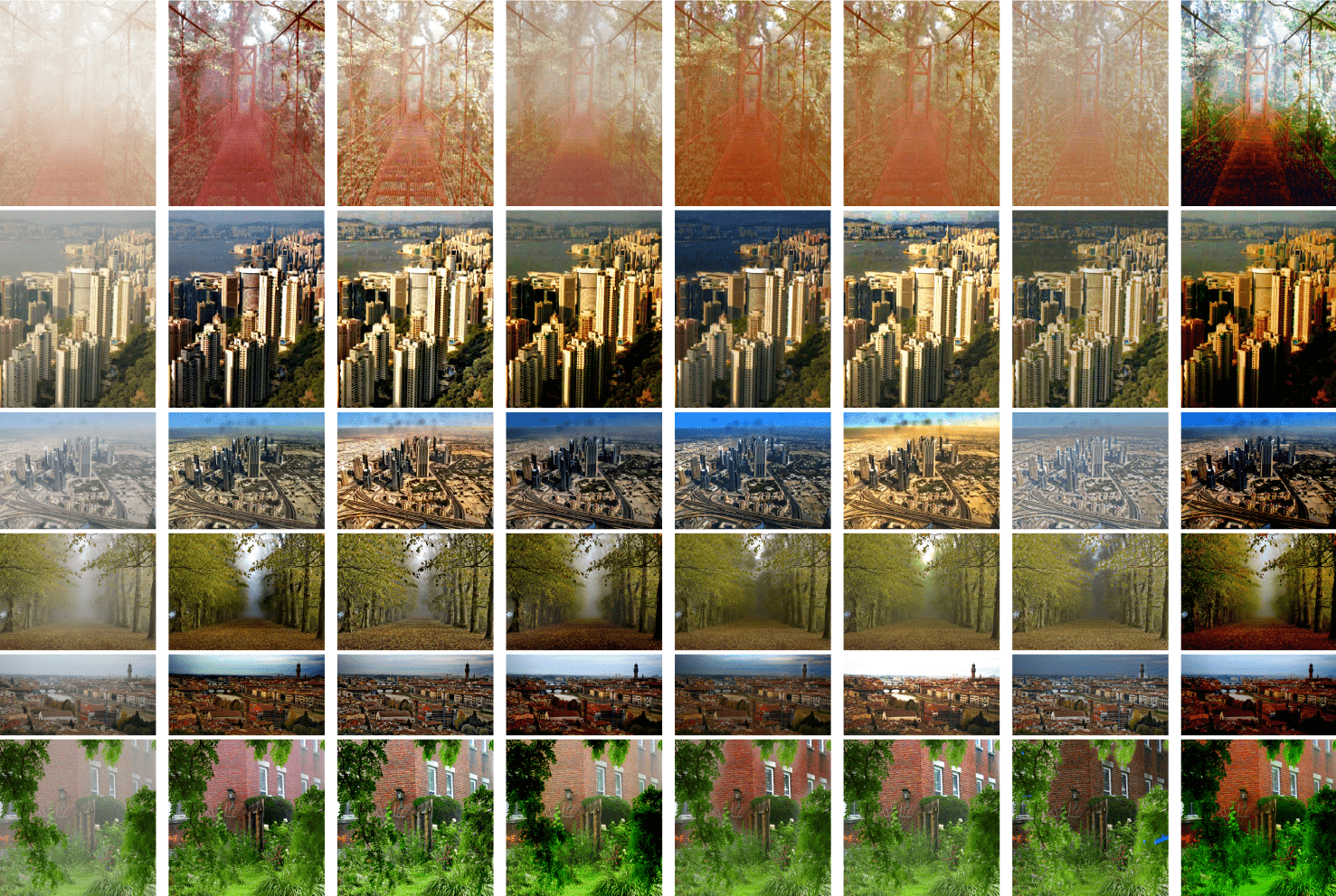}                 
                \caption{\textbf{Samples \#1 from our dataset with our methods compared to each other.} First column represents the original image taken in bad weather. The next three columns represent from left to right, the inverted intensities function for $\lambda$, the last three images use $\lambda = 0.35$. Each of these settings of $\lambda$ used ace\cite{Gatta:2002}, clahe\cite{Zuiderveld:1994} and stress \cite{Kolas:2011} respectively}
                \label{our_filterConf1}
  \end{center}%
\end{figure*}

We carry out our experiment on many images. We present here only the results of 6 images chosen in state-of-the-art database. We call these images: \textit{Bridge} for the image on the first line, \textit{Beach} for the second image, \textit{Dubai}, \textit{Forest}, \textit{Florence}, and \textit{House} for the third, fourth, fifth and sixth respectively.

For our experiment, we use the following six algorithms from advanced works in the literature: Berman \textit{et al.}\cite{Berman:2016}, Cho \textit{et al.}\cite{cho:2018}, Choi \textit{et al.}\cite{Choi:2015}, He \textit{et al.}\cite{He:2011}, Meng \textit{et al.}\cite{Meng:2013},  Tarel and Hauti{\`e}re \cite{Tarel:2009}. The configuration of these methods was done on the basis of the configurations used in the associated documents. Because of their implementation the case of the Meng \textit{et al.}\cite{Meng:2013} and Berman \textit{et al.}\cite{Berman:2016} methods can have various configurations. To stay clear and consistent about our work, for Meng, we have always used the clearest part of the sky or in the whole image. For Berman, we have chosen a single configuration present in the initial document.

\begin{figure*}[h!]
  \begin{center}
  \centering
  \includegraphics[width=0.99\textwidth]{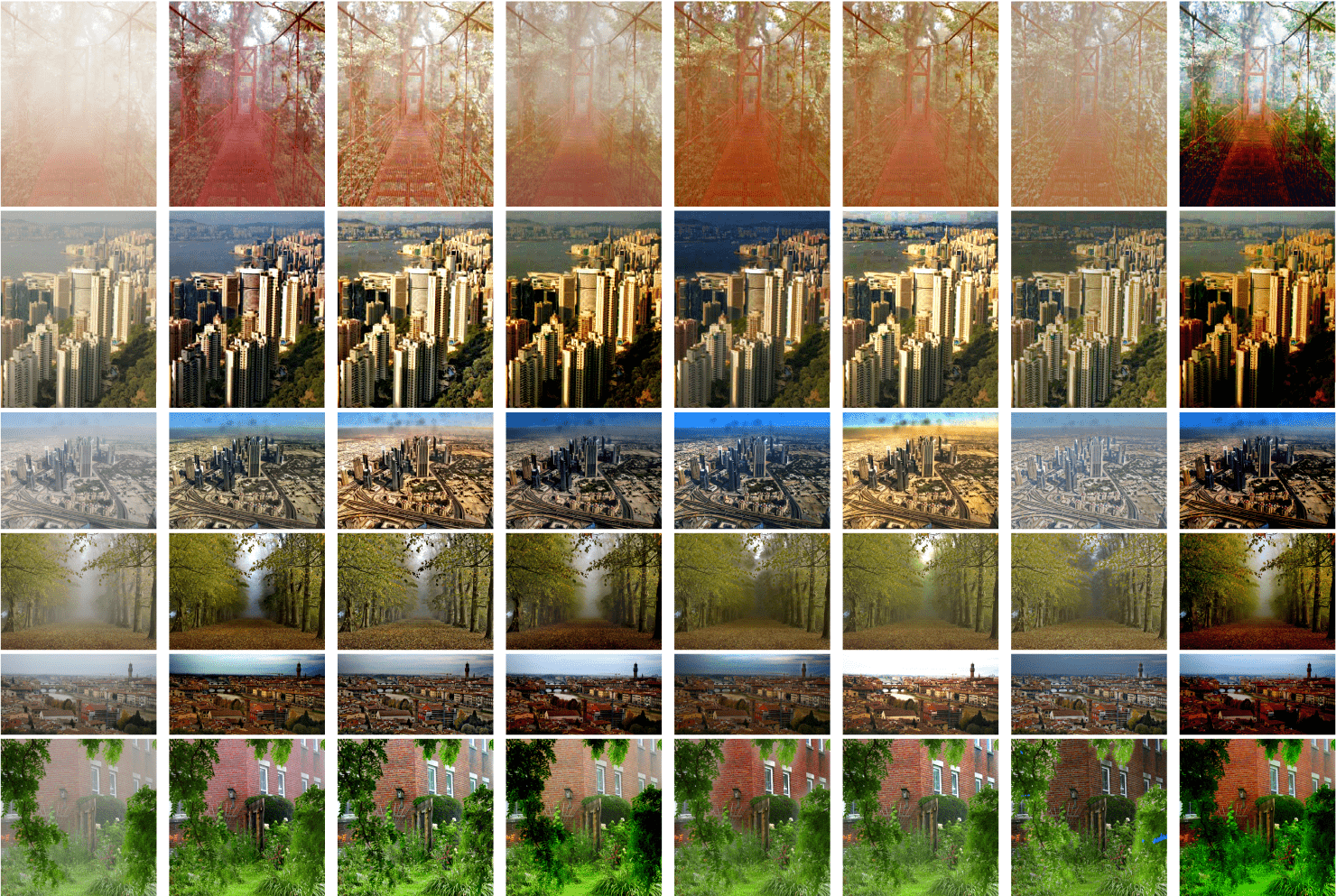}
                \caption{\textbf{Samples \#1 from our dataset with advanced techniques compared to our best configuration.} First column represents the original image taken in bad weather. The other columns represent from left to right, Berman \textit{et al.}\cite{Berman:2016}, Cho \textit{et al.}\cite{cho:2018}, Choi \textit{et al.}\cite{Choi:2015}, He \textit{et al.}\cite{He:2011}, Meng \textit{et al.}\cite{Meng:2013},  Tarel and Hauti{\`e}re \cite{Tarel:2009},  and Ours (the final best output selected from the previous figure) respectively.}
                \label{advancedConf1}
  \end{center}
\end{figure*}

For our family of algorithms, we consider two parameters, namely the inverse intensity function and the constant function $\lambda = 0.35$. We then applied these two parameters to three local contrast filtering algorithms, ace \cite{Gatta:2002}, clahe \cite{Zuiderveld:1994} and stress \cite{Kolas:2011}. This, therefore, results in six configurations for our family of algorithms. Fig. \ref {our_filterConf1} and Fig. \ref{advancedConf1} show the visual appearance of these compared configurations on the one hand, and on the other hand, of the best of these configurations with that of the advanced methods chosen. The complete settings of our procedure is given as supplementary material, although it is worth noting that we use the same settings for all the configurations presented. 

By analyzing the objective data from our family of algorithms in the Table. \ref{tab:table1_a}, we can notice there is little or no difference between \cite{Gatta:2002} and \cite{Kolas:2011} when we use our constant function $\lambda$ (columns 6 and 8). This is no longer true when we use a $\lambda$ function which is variable on space and on the three chromatic channels (columns 3 and 5). Even with the performance similarity noted for applying \cite{Gatta:2002} and \cite{Kolas:2011} when $\lambda$ is constant, their visual appearance is not identical. We invite the reader to zoom in to check the visual difference of these two renderings (column 5 and 7 in Fig. \ref{our_filterConf1}). This clearly suggests that the defogging measure used in \cite{Choi:2015} has certain drawbacks related to the similarity of appearance of haze-free images.

Table. \ref{tab:table1_a} of this set of images puts a bug in the ear that clahe surpasses the other local improvement techniques for the two configurations $\lambda$ used. We also invite readers here and elsewhere to zoom in to analyze the visual difference between the methods presented, even if it is a completely subjective task. 

As said above, Table. \ref{tab:table1_b} shows the results of state-of-the-art methods compared to ours. In the first line which corresponds to the \textit{Bridge} image, our best configuration using stress outperforms state-of-the-art approaches in terms of objective evaluation. The second-best output is objectively  given by  He \textit{et al.}\cite{He:2011} method, while the third-best is obtained with Meng \textit{et al.}\cite{Meng:2013} approach.  
\begin{table*}
  \centering
\begin{tabular}[t]{c c c c c c c c c c c c}
\toprule
   image   & original & ace(2) & clahe(2) & stress(2)  & ace(1) & clahe(1) & stress(1) \\
\midrule
\it{{Bridge}}   &  2.451 & 0.370 & 0.329 & 0.239 & 0.487 & 0.275   & 0.487\\
\it{{Beach}}& 0.580 & 0.382 & 0.168 & 0.172 & 0.353 & 0.234 &  0.353\\
\it{{Dubai}} & 1.036 & 0.253 & 0.135 & 0.139 & 0.336 & 0.303 & 0.335\\
\it{{Forest}} & 0.389 & 0.176 &  0.118 &  0.131 & 0.257 & 0.178 & 0.257\\
\it{{Florence}} & 0.847 & 0.438 & 0.197 & 0.244 & 0.441 & 0.310 & 0.441\\
\it{{House}} & 0.261 & 0.154 & 0.119 & 0.122 & 0.201 & 0.160 & 0.201\\ 
\bottomrule
\end{tabular}
\caption{\footnotesize Our method compared to each other . }
\label{tab:table1_a}
\end{table*}

In the image \textit{Beach}, the best score is given with the Choi \textit{et al.} \cite{Choi:2015} method, and the second place is deserved by Cho \textit{et al.} \cite{cho:2018}, while our clahe-based method won the third place.

Like the first series of comparison, our method leads the ranking with \textit{Dubai} image. Here, the approaches Choi \textit{et al.}\cite{Choi:2015} and Cho \textit{et al.}\cite{cho:2018} occupy second and third places respectively.

In the image \textit{Forest} , we got the second-best place, while in the image \textit{Florence}, we are again third.  We are also third  with \textit{House}, however Tarel and Hauti{\`e}re \cite{Tarel:2009} have an identical score. Except for the first image, all of our best results used clahe.

The first place in \textit{Forest} is won with Cho \textit{et al.}\cite{cho:2018} approach, and the third is awarded to Berman \textit{et al.}\cite{Berman:2016}. For  the case of image \textit{Florence}, Berman \textit{et al.}\cite{Berman:2016} and Cho \textit{et al.}\cite{cho:2018} are respectively first and second. In \textit{House}, the first and second place are respectively occupied by Cho \textit{et al.}\cite{cho:2018}, Choi \textit{et al.}\cite{Choi:2015} methods.

We refer the reader to the additional documents provided for further comparison and analysis of the proposed method.

\begin{table*}
  \centering
\begin{tabular}[t]{c c c c c c c c c c c c}
\toprule
   image   & original & Berman & Cho & Choi  & He & Meng & Tarel & Ours \\
\midrule
\it{{Bridge}}   &  2.451 & 0.393 & 0.414 & 0.775 & 0.295 & 0.345   & 0.617 & 0.239  \\
\it{{Beach}} & 0.580 & 0.176 & 0.161 & 0.154 & 0.204 & 0.225 & 0.212 & 0.168 \\
\it{{Dubai}} & 1.036 & 0.230 & 0.163 & 0.154 & 0.184 & 0.193 & 0.443 & 0.135\\
\it{{Forest}} & 0.389 & 0.128 & 0.114 & 0.135 & 0.166 & 0.178 & 0.182 & 0.118\\
\it{{Florence}} & 0.847 & 0.155 & 0.158 & 0.200 & 0.235 & 0.646 & 0.226 & 0.197\\
\it{{House}} & 0.261 & 0.135 & 0.112 & 0.116 & 0.133 & 0.132 & 0.119 &  0.119\\
\bottomrule
\end{tabular}
\caption{\footnotesize Our method compared to Berman \textit{et al.} \cite{Berman:2016}, Cho \textit{et al.}\cite{cho:2018}, Choi \textit{et al.} \cite{Choi:2015}, He \textit{et al.}\cite{He:2011}, Meng \textit{et al.} \cite{Meng:2013}, Tarel and Hauti{\`e}re \cite{Tarel:2009},  and Ours (Best)}
\label{tab:table1_b}
\end{table*}

We can notice here that our clahe-based methods appear in the Table. \ref{tab:table1_a} in second and third place with a $lambda$ constant and a non-constant functions respectively for the image \textit{Bridge}. This means that if we do not mention stress method for these series, our technique based on clahe will remain in the three main methods for this image. Thus,  it will also appear in the three leading methods in Table. \ref{tab:table1_b}.

To sum up the results in the Table. \ref{tab:table1_b}, our method clearly appears in the three leading methods  in this comparison set. The other remark for these image samples is the stability of the method of Cho \textit{et al.}\cite{cho:2018} which also appears in the three leading results except for the first series of images. Other stable techniques are the methods of Choi \textit{et al.} \cite{Choi:2015} and Berman \textit{et al.} \cite{Berman:2016} which appear quite often in dominant results for this set of images.

\section{Discussion}
From this experiment and the results obtained, we think that the algorithm obtains its dehazing property from the first filter in the $\lambda$ function  and from the local color contrast algorithm used. We believe this will open the door to further exploration of such a dehazing algorithm family in the community. Even if the two functions $\lambda$ presented are not very successful with a dense haze case, we think that by finding appropriate $\lambda$ and a good local contrast filter, we can diffuse these extreme cases which are more suited to deep learning algorithms to date. Examples of such algorithms can be found in \cite{Ancuti:2019}.

Another interesting point is that all these algorithms, that is to say, ours and state-of-the-art approaches can be tuned. Here we only set a default configuration to work with. This suggests that these algorithms cannot be strictly  evaluated on only these default parameters and that our comments are only related to the chosen settings.

\section{Conclusion}
In this paper, we present a general-purpose framework that uses local contrast techniques to enhance the images taken in foggy conditions. Even with its simplicity, our algorithm is fast, it  has no post-processing trick, it clearly competes with advanced work, and it shows promising ways of exploring dehazing algorithms. As many non-learning dehazing procedures, its main drawback is that it fails to enhance dense haze, fog, mist, smoke situation. In future work, we will explore this  engaging problem within this family of dehazing algorithms.

%
%
\bibliographystyle{splncs04}
\bibliography{egbib}
\appendix[\textbf{Supplementary Materials}]

The following items are provided in the supplementary material:
  \begin{itemize}
  \item Relationship between ace, clahe and stress
  \item What properties of contrast enhancement algorithms are empirically necessary for dehazing in our framework?
  \item Adding prior and optmization procedure to our algorithm
  \item Experimental setup details
  \item More comparison and analyses
  \item Code
  \item Miscellaneous
  \end{itemize}

\section{Relationship between ace, clahe and stress}
The general idea behind \textit{ the local contrast improvement} algorithm is that a \textbf{weighted average} of the pixel intensities controlled by a \textbf{standard deviation} is calculated in a \textbf{contextual region}  of the pixel of interest based on a given \textbf{sampling scheme}. This computation is made in  the spatial domain (e.g., ace \cite{Gatta:2002}, \cite{Rizzi:2003}, clahe \cite{Zuiderveld:1994}, stress \cite{Kolas:2011}). Each of these filtering methods is a form of unsharp masking technique  in which high frequency components have remained dominant since unsharp masking presuppose blending an image's high frequency components and low frequency components to improve its quality.

\section{What properties of contrast enhancement algorithms are empirically necessary for dehazing in our framework?}

In the main paper, we define two filters that help us  to carry out the dehazing task in our procedure. The first filter is defined as follows:
\begin{eqnarray}
  \label{first_filter}
  f_{init} (\mathbf{x}_i) =  \frac{\mathbf{x}_i - \mathbf{x}_{min}}{\mathbf{x}_{max} - \mathbf{x}_{min}}
\end{eqnarray}

We then further extend this filter to the more general filter as the following:
\begin{eqnarray}
  \label{general_filter}
  f_g (\mathbf{x}_i) = f_{init} (\mathbf{x}_i) - \lambda(\mathbf{x}_i)\lVert f_{init} (\mathbf{x}_i) \rVert
\end{eqnarray}

The back-end and final filter  is an image local contrast enhancement scheme that takes the output of the front-end filter and it can be expressed as follows:
\begin{eqnarray}
  \label{last_filter}
  f_{final} (\mathbf{x}_i) = f_{LCE}(f_g (\mathbf{x}_i) )
\end{eqnarray}

Consequently, our solution consists, roughly speaking, in applying a normalization scheme to the entire image by blackening pixels, then using a technique of enhancement of the low-contrast generated to improve the image in order to obtain the final haze-free image.

In our experiment, we use three \textit{spatial local contrast enhancement} algorithms. However, Ace \cite{Gatta:2002}, \cite{Rizzi:2003}, clahe \cite{Zuiderveld:1994} and stress \cite{Kolas:2011} are not only local contrast enhancement methods, but they are also low-contrast enhancement techniques.  Thus, one can wonder which  criterion (or criteria) shares these algorithms so that they contribute to improve the image taken in foggy conditions.

To check out which criteria help in the defogging procedure, we empirically begin by analyzing the output of Equations \ref{general_filter} and \ref{last_filter} separately.
\begin{figure}[ht]
  \begin{center}
  \centering
  \includegraphics[width=.99\textwidth]{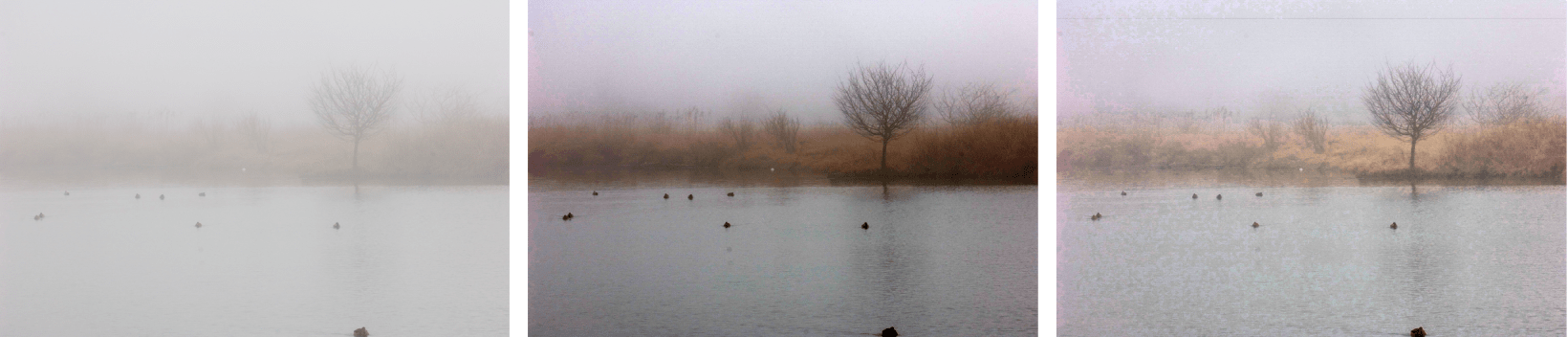}
  \caption{\textbf{An Example of outputs  from the two filters:} Lack in Desert image on hazy condition (left), and output images of the  two  filters (middle \& right). The one in the middle is the result of the first filter in Equation \ref{general_filter} ($\lambda$ here represents the inverted intensites function), while the other use clahe \cite{Zuiderveld:1994}  as the filter $f_{LCE}$ in Equation \ref{last_filter} for its output.}
  \label{First_example}
  \end{center}
\end{figure}

Looking closely at Fig. \ref{First_example}, we can notice that the haze density seems uniform throughout the image and  that the first filter not only removes the fog in the image, but it also darkens the scene. In the last image (from left to right) which represents the last filter, the low-contrast observed in the middle image is restored. This makes sense since these local contrast enhancement techniques also enhance a low-contrast image. 

\begin{figure}[ht]
  \begin{center}
  \centering
  \includegraphics[width=.99\textwidth]{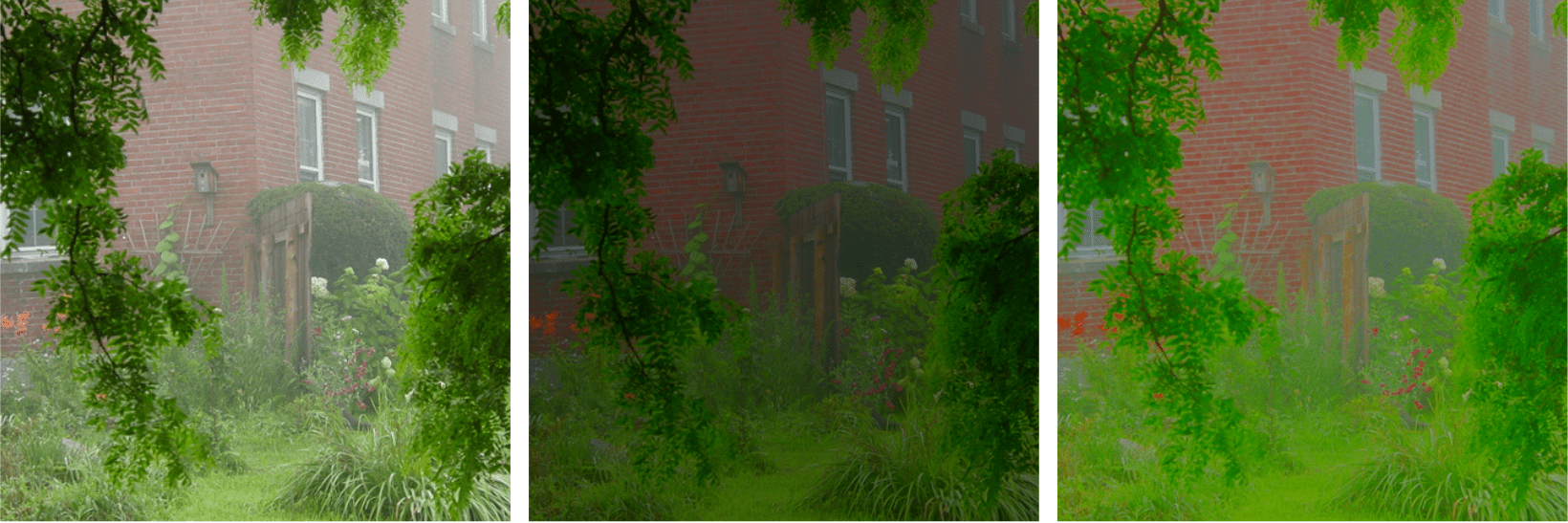}
  \caption{\textbf{An Example of outputs  from the two filters using gamma correction as the last filter} House image on hazy condition (left), and output images of the  two  filters (middle \& right). The one in the middle is the result of the first filter in Equation \ref{general_filter} ($\lambda=0.35$ here one can observe that the algorithm homogenizes and blackens the original somewhow), while the other use gamma correction  as the filter $f_{LCE}$ in Equation \ref{last_filter} for its output. This example clearly shows that gamma correction does not help for removing fog}
  \label{Second_example}
  \end{center}%
\end{figure}

At the same time, applying gamma correction to the output of the Equation \ref{general_filter} using a constant $\lambda$ function does not dehaze the image that much. An example is given in Fig. \ref{Second_example}.

For our investigation, We  divide filters into three groups:
\begin{enumerate}
\item The first filter in Equation \ref{general_filter}: As we said in the main paper, this filter is already a dehazing filter on its own when the inverted intensity function is used. As we observe that the dehazing procedure is more efficient with the inverted intensity function than with the constant function $\lambda$, we consider only the first function $\lambda$  in our comparison.
\item The second filter: We consider the function $f_{LCE}$ as a  low-contrast enhancement method that has a  local contrast enhancement property such as ace\cite{Gatta:2002}, clahe\cite{Zuiderveld:1994} and stress \cite{Kolas:2011}
  \item The second option for $f_{LCE}$: A low-contrast enhancement algorithm that does not incorporate a local contrast requirements, for example, simple histogram equalization, gamma correction or any other stationary contrast mapping function.
\end{enumerate}

After having observed more than $200$ images with these filters on their own, the following observations turn up. The first filter emerges as a more powerful dehazing procedure, but it often dims the scene. The second pops up as a more likely dehazing filter  on objects close to the camera. We notice that both filters remove fog from the original image and that their order (when combined) also has its impact. The best combination in our opinion is the one indicated in our procedure above. Fig. \ref{First_example} shows the outputs of the filter in Equation \ref{general_filter} (without any local contrast enhancement technique) and Equation \ref{last_filter} (using clahe as a local contrast enhancement procedure) respectively.

While we are testing the effect of filters that have low-contrast enhancement property but do not have local contrast enhancement property (e.g., histogram equalization), we notice that the haze suppression property is not meaningful with these filters as is the case of the previous two filters.

To recap, we empirically note that the first two filter groups have somewhat improved  the hazy image on their own, and the third does not significantly enhance the foggy image by its own nature. However, we need to tone down our conclusion here. Vanilla histogram equalization, even without local contrast enhancement property, is more suited to dehazing than to a simple gamma correction. This suggests the low-contrast property is not the only property that works for histogram equalization and that it benefits some degree of contrast enhancement properties that are suitable with dehazing. Another critical point for dehazing with histogram equalilization and the gamma correction filtering procedure is the increase in the density of haze related  to the depth. Since these algorithms are global contrast improvement techniques, they are clearly less effective than local contrast filtering techniques which acclimate to the local context of images as  the human visual system does \cite{Land:1971}, \cite{Pizer:1987}.

Reasons why we do not mention that the framework can work with any kind of contrast enhancement technique are already justified above. Furthermore, we observe difference of the rendering of a basic implementation of histogram equalization algorithm depending on the library used, and the dehazing property of this filter is not always obvious. So from our experiment, both low-contrast and local contrast improvement properties (e.g., \cite{Harris:1977}, \cite{Narendra:1981}, \cite{Gatta:2002}, \cite{Rizzi:2003}, \cite{Kolas:2011},  \cite{Zuiderveld:1994}) are necessary for the image enhancement algorithm  in the dehazing procedure for general cases of the definition of Equation \ref{general_filter}.

\section{Adding prior and optmization procedure to our algorithm}
\label{optim}
The solution propose in the main document using the inverted intensity function tends to darken too much the image.
To overcome this, we assume our haze solution has a white balancing  issue at some extent. Typically, our inverted intensity function resemble to the transmission map estimation used in \cite{He:2009}. So we applied this prior to our proposed solution and we obtain the following definition:
\begin{equation}
  \label{eq:04}
  \lambda(\mathbf{x}_i) = 1 - \min_{c}(\min_{\mathbf{y}_i \in \Omega}(\frac{f_{init} (\mathbf{y}_i)}{A^c}))
\end{equation}

where $A^c$ is computed as the atmospheric light of the image $f_{init}$. We transpose the problem in hands as a white balancing problem, and we use soft matting \cite{Levin:2006}, \cite{Hsu:2008} to refine our transmission  $\lambda$ by minimizing the following cost function:

\begin{equation}
  \label{eq:04}
  E(\lambda) = \lambda^T L \lambda + \beta(\lambda - \hat{\lambda})^T(\lambda - \hat{\lambda})
\end{equation}

The optimal $\lambda$ is  found  by solving the sparse linear system defined as follows:
\begin{equation}
  \label{eq:04}
   (L  + \beta U)\lambda = \beta \hat{\lambda}
\end{equation}

Once our optimal $\lambda$ obtained from the above equation, we normalize it using $f_{init}$ and we call the result $\lambda_{norm}$. We then compute its inverted intensity $\lambda_{invert} = 1 - \lambda_{norm}$. The final matrix $\lambda_{final}$ is computed as the minimum of $\lambda_{norm}$ and $\lambda_{invert}$. This expression is put in Equation \ref{general_filter} to deduce the dehazed result in Equation \ref{last_filter}.

We can observe here that we are not  only using the exact definition of the transmission employed in \cite{He:2009}. In fact, the exact definition of the transmission of \cite{He:2009} does not provide satisfactory results. The resulting image contains an unreasonable number of visible artifacts. In order to decrease the number of artifacts, we use two normalized functions from which we consider the minimum values to form the final function. This new approach gives a more satisfactory result. There are certainly other ways to improve the rendering using the transmission described in \cite{He:2009}, but here our objective is to show how we can use this transmission in our solution or to give usable tracks for future research.
\begin{figure}[ht]
  \begin{center}
    \centering
    \includegraphics[width=.99\textwidth]{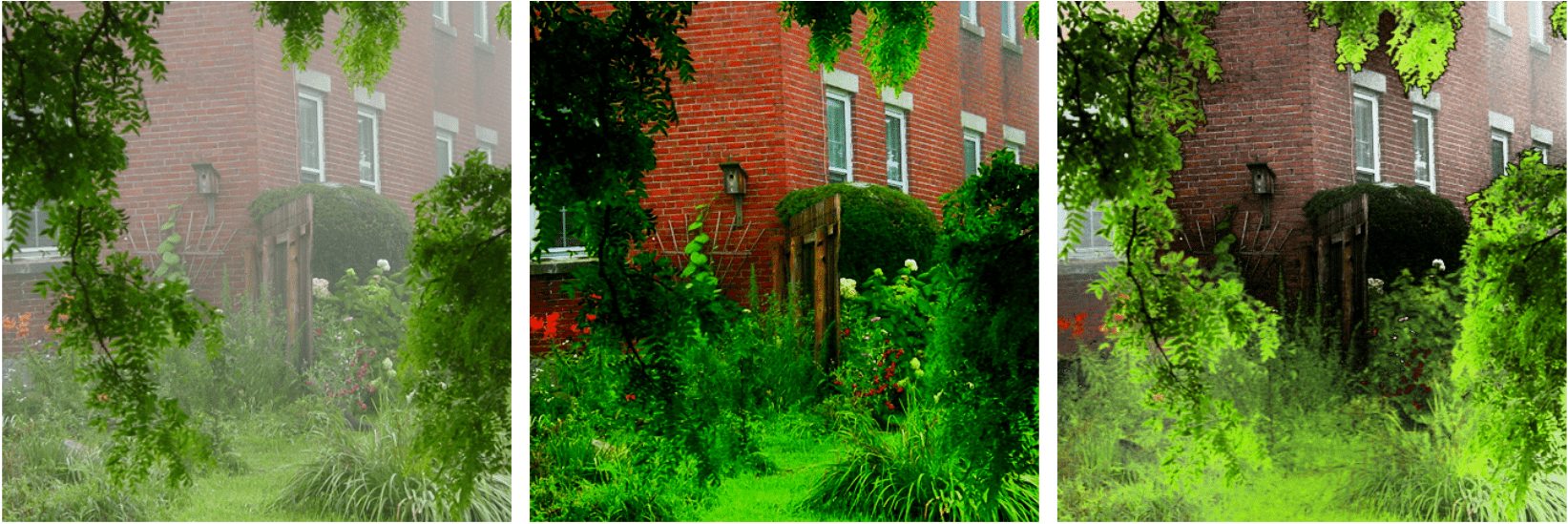}
   \caption{\textbf{Outputs  from the two filters using our inverted intensities map and the transmission function defined in \cite{He:2009}} House image on hazy condition (left), and output images of two functions $\lambda$ (middle \& right). The one in the middle is the result of the second filter in Equation \ref{last_filter} (with $\lambda= 1 - f_{init}$),  the other uses $\lambda$ as the transmission map in Section \ref{optim}. One can observe that the haze removal is effective with this last approach, however the result may contain some artifacts.}
  \end{center}%
  \label{Third_example}
\end{figure}

The Figure above shows that adding more complexity to our haze removal procedure does not necessarily improve previous results. We  have indeed less darkness than expected, though some parts of the image appear unusual. This section suggests that the simplicity of the proposed algorithm makes it easier to add more complexity; as we can see we are augmenting it here with a physical formulation.

\section{Experimental setup details}

The main procedure is quite simple, and it can be implemented using openCV, Matlab, Scikit-image, openGL or any other computer vision framework. The local contrast enhancement algorithms ace \cite{Gatta:2002}, \cite{Rizzi:2003}, stress \cite{Kolas:2011} or clahe \cite{Zuiderveld:1994} are tuned as follows:

\begin{enumerate}
\item ace: We use the fast algorithm described in \cite{Getreuer:2012} for fast computations. We set two parameters from the previous paper:
  \begin{enumerate}
  \item $\alpha$ that defined the slope function is set to $5$
    \item $w=1/\sqrt{x^2 +y^2}$ the weighted function which use the level interpolation formulation with $J = 8$ levels.
  \end{enumerate}
\item clahe: Here we use two settings depending on the fact that we are working with  $\lambda=0.35$ or with $\lambda=1 - f_{init}$. We essentially use the \textit{clip limit} and the \textit{kernel size} as the main parameters of the clahe algorithm. For both configurations, we use the default value of $0.008$ for the \textit{clip limit}
  \begin{enumerate}
  \item $\lambda=0.35$: we set the \textit{kernel size} as the product of $1/8 width \times 1/8 height$ of the image
    \item $\lambda=1 - f_{init}$: We use the default value $800$ for all the images.
  \end{enumerate}
\item stress: We use the following parameters from the initial framework:
  \begin{enumerate}
  \item the number of samples $n_s=5$
  \item the number of iteration $n_i = 150$
  \item the spray radius  is set to the maximum of the width and height of the image.
  \end{enumerate}
\end{enumerate}
For state-of-the-art dehazing algorithms used in our experiment, we use the default parameters tuning described in the original paper except for Berman \textit{et al.}\cite{Berman:2016} and Meng \textit{et al.}\cite{Meng:2013}. For Meng \textit{et al.}\cite{Meng:2013}, we have always used the clearest part of the sky or at a place where the selection is made on a white object in the scene. Whereas for Berman \textit{et al.}\cite{Berman:2016} we choose the following parameters ﻿$\gamma = 1$, and the atmospheric light $A=[0.81, 0.81, 0.82]^T$ for all images.

\section{More comparison and analyses}
In this Section, we  first report objective measure (Table \ref{tab:table1_a} and Table \ref{tab:table1_b}) of the experiment made in the main paper based on the metrics developed in \cite{Hautiere:2008}. In the Table \ref{tab:table1_a} and the Table\ref{tab:table1_b}, the measures $ e $ and $\bar{r}$ must be high while the measure $\Sigma$ must be low in terms evaluation of fog suppression algorithms. On this basis, we clearly see that the results obtained by our approach compete with advanced algorithms.

We then effectuate more visual inspection of the results of our three local color
contrast enhancement algorithm in our framework that we compare to state-of-the-art methods. Here, because our analysis is based on a suggestive approach, we do not present the six configurations as was the case with the main document. We alternate the presentation of the renderings according to the two lambda functions used to which we compare four algorithms among the six advanced algorithms previously used in the main document.

In this document, rather than presenting a large number of compressed results whose quality cannot really be analyzed by the reader, we present a limited number of our results which can be analyzed by the reader more easily.
We will put all of our data online very soon so that it is available to the community.

\begin{table*}
\begin{tabular}[t]{c c c c c c c c}
  \toprule
  image &   metrics  & Ace(2) & clahe(2) & stress(2) & Ace(1) & clahe(1) & stress(1)  \\
  \midrule
  & e &      1.5004e+03 & 1.3976e+03 & 1.4510e+03 & 1.2452e+03 & 1.4872e+03 & 1.3817e+03 \\
&  $\Sigma$ &  1.9903e-04 & 0.2152 & 0.0184 & 1.0663e-05
 & 0.0796  & 0.0141 \\
&  $\bar{r}$ & 6.4746 & 5.4336 & 8.3459 & 4.3654 & 7.4685 & 14.1843\\
  \midrule   
& e &       0.0749 & 0.0502 & 0.2896 & 0.0587 & 0.0995 & 0.2442 \\
&  $\Sigma$ &  3.2257e-05 & 0.3222 & 0.0974 & 1.2096e-05
 & 0.0142  & 0.0205 \\
  &  $\bar{r}$ & 1.6979 & 1.3737 & 1.3587 & 1.2503 & 1.5416 & 1.7714\\
\midrule   
& e &       0.3007 & 0.3886 & 0.4330 & 0.3336 & 0.2670 & 0.5765 \\
&  $\Sigma$ &  0.2524 & 0.3222 & 0.2093 & 1.3748e-05
 & 6.2425e-05  & 0.0055 \\
&  $\bar{r}$ & 3.1414 & 2.0935 & 2.8515 & 1.8727 & 1.9007 & 2.8913\\
\midrule   
& e & 0.3255 & 0.2544 & 0.3057 & 0.1885 & 0.2559 & 0.3231 \\
&  $\Sigma$ &  1.5089e-04 & 0.3553 & 0.0882 & 4.4929e-05
 & 0.0648  & 0.0631 \\
&  $\bar{r}$ & 2.4058 &  1.6371 & 2.6854 & 1.6429 & 1.7346 & 3.1291\\
\midrule   
& e & -0.0917 & -0.1293 & 0.0050 & 0.0692 & 0.0747 & 0.2295 \\
&  $\Sigma$ &  1.3599e-05 & 0.0730 & 0.0882 & 5.7027e-06
 & 0.0125  & 0.0204 \\
&  $\bar{r}$ & 2.2124 &  1.6648 & 1.7559 & 1.9170 & 2.2655 & 2.9467\\
\midrule   
& e & 0.0069 & 0.0306 & 0.2814 & 0.1016 & 0.1982 & 0.1656 \\
&  $\Sigma$ &  6.3973e-05 & 0.3961 & 0.2814 & 3.3670e-05
 & 0.1189  & 0.0529 \\
  &  $\bar{r}$ & 2.4009 &  1.4403 & 1.4572 & 1.5347 & 1.5691 & 2.0522\\
\midrule
\bottomrule
\end{tabular}
\caption{\footnotesize Our method compared to Berman \textit{et al.} \cite{Berman:2016}, Cho \textit{et al.}\cite{cho:2018}, Choi \textit{et al.} \cite{Choi:2015}, Fattal \cite{Fattal:2014}, He \textit{et al.}\cite{He:2011}, Meng \textit{et al.} \cite{Meng:2013}, Tarel and Hauti{\`e}re \cite{Tarel:2009}, Ours (Automatic STRESS) and ours(Fast)}
\label{tab:table1_a}
\end{table*}


\begin{table*}
\begin{tabular}[t]{c c c c c c c c}
\toprule
image &   metrics  & Berman & Cho & Choi & He & Meng & Tarel  \\
\midrule
& e &       1.0752e+03 & 1.1079e+03 & 163.4217 & 691.7711 & 534.3373 & 197.2410 \\
&  $\Sigma$ &  0.0034 & 2.3813e-04 & 0 & 1.3269e-04
 & 0.0017  & 0 \\
&  $\bar{r}$ & 5.0297 & 5.5577 & 2.3449 & 4.0315 & 4.1035 & 2.7611\\
\midrule
& e &       0.0783 & 0.1157 & 0.1361 & 0.2714 & 0.1593 & 0.3936 \\
&  $\Sigma$ &  0.0810 & 0.0870 & 0.1860 & 0.0023
 & 0.0024  & 0 \\
&  $\bar{r}$ & 1.5234 & 2.0351 & 1.5672 & 1.3563 & 1.9339 & 1.6603\\
\midrule
& e &       0.3428 & 0.5221 & 0.4975 &  0.6521 & 0.4822 & 0.4745 \\
&  $\Sigma$ &  0.0035 & 0.0380 & 0.0013 & 0.0023
 & 0.0014  & 0 \\
&  $\bar{r}$ & 2.0490 & 3.0302 & 2.0163 & 2.2240 & 2.6319 & 2.1493\\
\midrule
& e &      0.1833 & 0.1710 & 0.2666 &  0.3104 & 0.2416 & 0.5729 \\
&  $\Sigma$ &  0.1029 & 0.0835 & 0.0793 & 0.0065
 & 0.0040  & 2.5431e-04 \\
&  $\bar{r}$ & 1.9830 & 2.7679 & 1.7357 & 1.3044 & 1.3818 & 1.5025\\
\midrule
& e &      0.1658 & 0.1205 & -0.0212 &  0.3886 & 0.0381 & 0.4568 \\
&  $\Sigma$ &  0.1092 & 0.0896 & 0.1428 & 0.0043
 & 6.3168e-04  & 0 \\
&  $\bar{r}$ & 2.0207 & 2.3525 & 1.5966 & 1.8087 & 1.6475 & 2.4840\\
\midrule
& e &      0.0519 & 0.0362 & 0.1060 &  0.1958 & 0.2013 & 0.2319 \\
&  $\Sigma$ &  0.0861 & 0.1076 & 0.2078 & 0.0269
 & 0.0464  & 0.0337 \\
&  $\bar{r}$ & 1.6679 & 2.3421 & 1.4748 & 1.4209 & 1.6516 & 1.7850\\
\midrule
\bottomrule
\end{tabular}
\caption{\footnotesize Our method compared to Berman \textit{et al.} \cite{Berman:2016}, Cho \textit{et al.}\cite{cho:2018}, Choi \textit{et al.} \cite{Choi:2015}, Fattal \cite{Fattal:2014}, He \textit{et al.}\cite{He:2011}, Meng \textit{et al.} \cite{Meng:2013}, Tarel and Hauti{\`e}re \cite{Tarel:2009}, Ours (Automatic STRESS) and ours(Fast)}
\label{tab:table1_b}
\end{table*}

\begin{figure*}[h!]
  \begin{center}
        \centering
                \includegraphics[width=0.99\textwidth]{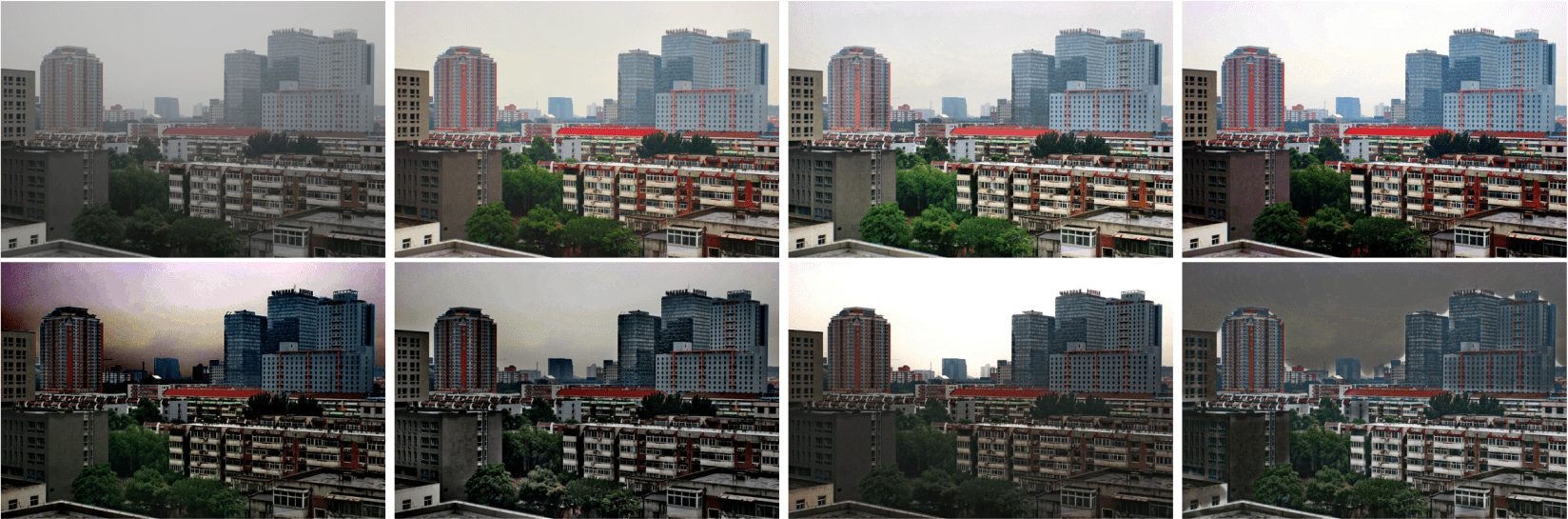}          
\caption{\textbf{Samples \#2 from our dataset with our methods compared to each other and to state-of-the-art methods.} First column represents the original image taken in bad weather. The next three columns represent from left to right, settings of $\lambda=0.35$ using ace\cite{Gatta:2002}, clahe\cite{Zuiderveld:1994} and stress \cite{Kolas:2011} respectively. Bottom row with Berman \textit{et al.}\cite{Berman:2016}, Cho \textit{et al.}\cite{cho:2018}, Meng \textit{et al.}\cite{Meng:2013} and Tarel and Hauti{\`e}re \cite{Tarel:2009}}
\label{our_filterConf_2}
\end{center}
\end{figure*}

\begin{figure*}[h!]
  \begin{center}
        \centering
                \includegraphics[width=.99\textwidth]{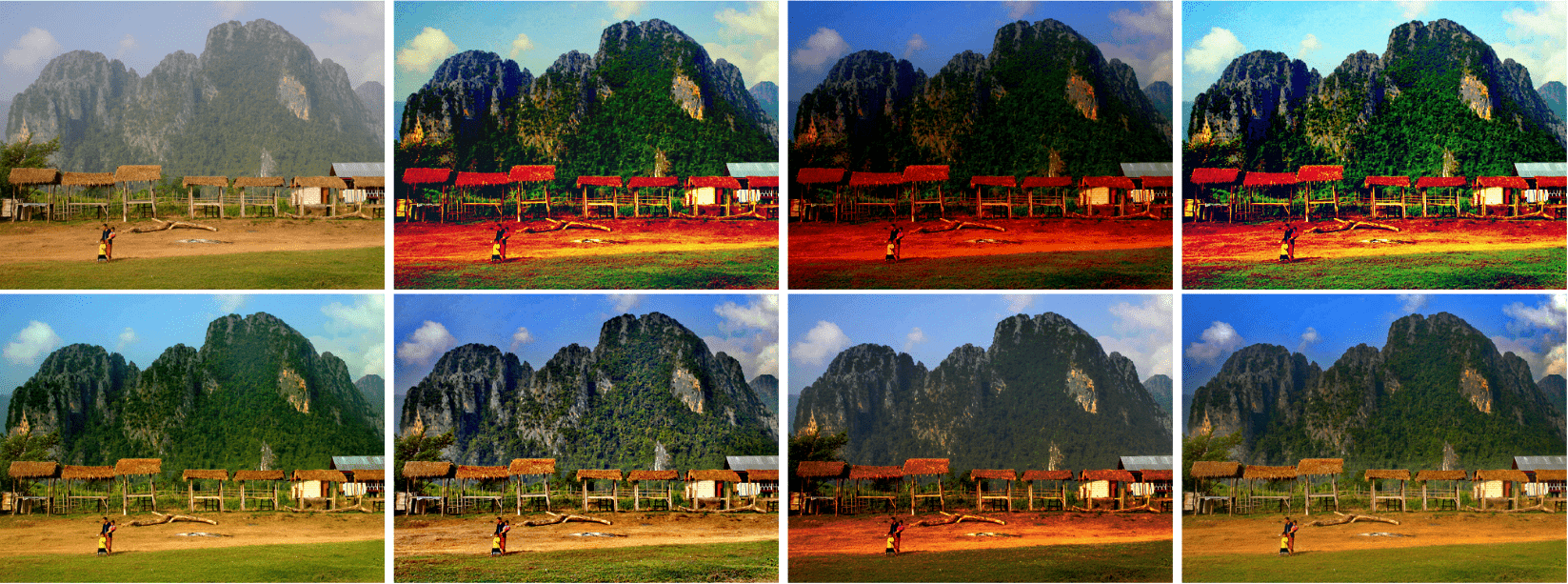}      
\caption{\textbf{Samples \#8 from our dataset with our methods compared to each other and to state-of-the-art methods.} First column represents the original image taken in bad weather. The next three columns represent from left to right, settings of $\lambda=1 - f_{init}$ using ace\cite{Gatta:2002}, clahe\cite{Zuiderveld:1994} and stress \cite{Kolas:2011} respectively. Bottom row with Berman \textit{et al.}\cite{Berman:2016}, Cho \textit{et al.}\cite{cho:2018}, Choi \textit{et al.}\cite{Choi:2015} and He \textit{et al.}\cite{He:2011}}
\label{our_filterConf13}
\end{center}
\end{figure*}

\begin{figure*}[h!]
  \begin{center}
        \centering
                \includegraphics[width=0.99\textwidth]{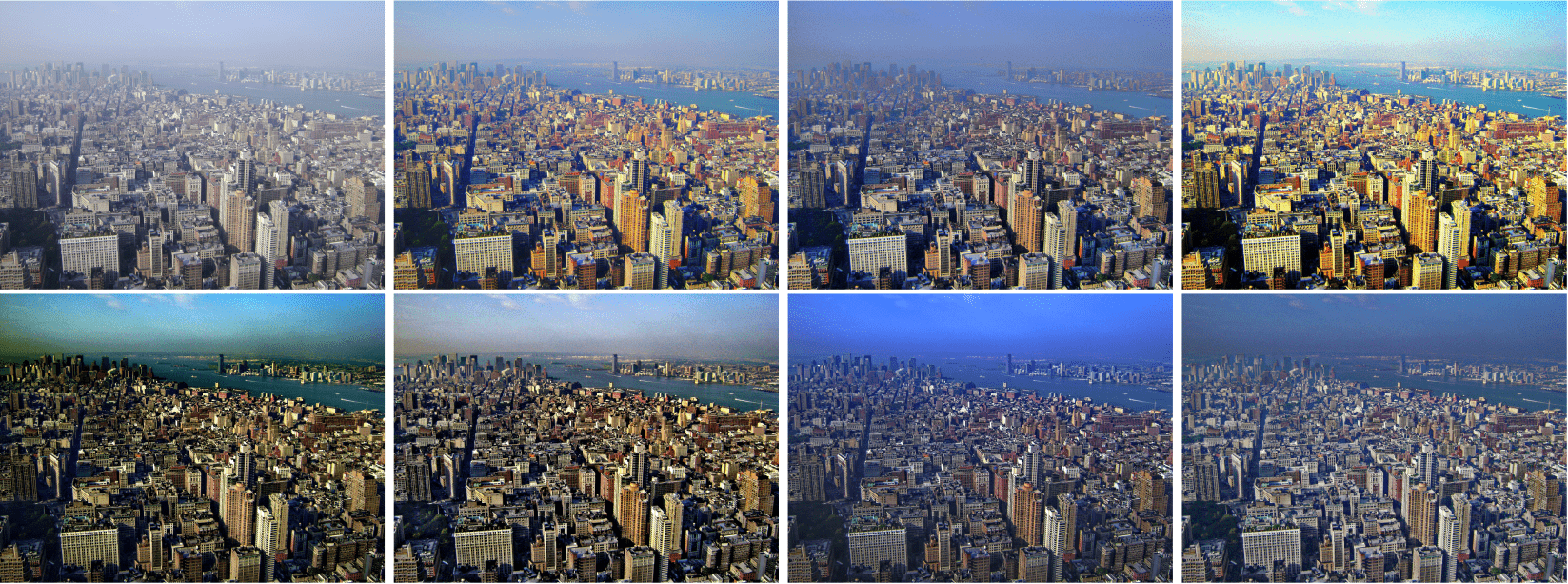}  
\caption{\textbf{Samples \#9 from our dataset with our methods compared to each other and to state-of-the-art methods.} First column represents the original image taken in bad weather. The next three columns represent from left to right, settings of $\lambda=0.35$ using ace\cite{Gatta:2002}, clahe\cite{Zuiderveld:1994} and stress \cite{Kolas:2011} respectively. Bottom row with Berman \textit{et al.}\cite{Berman:2016}, Cho \textit{et al.}\cite{cho:2018}, Meng \textit{et al.}\cite{Meng:2013} and Tarel and Hauti{\`e}re \cite{Tarel:2009}}
\label{our_filterConf13}
\end{center}
\end{figure*}

\begin{figure*}[h!]
  \begin{center}
        \centering
                \includegraphics[width=.99\textwidth]{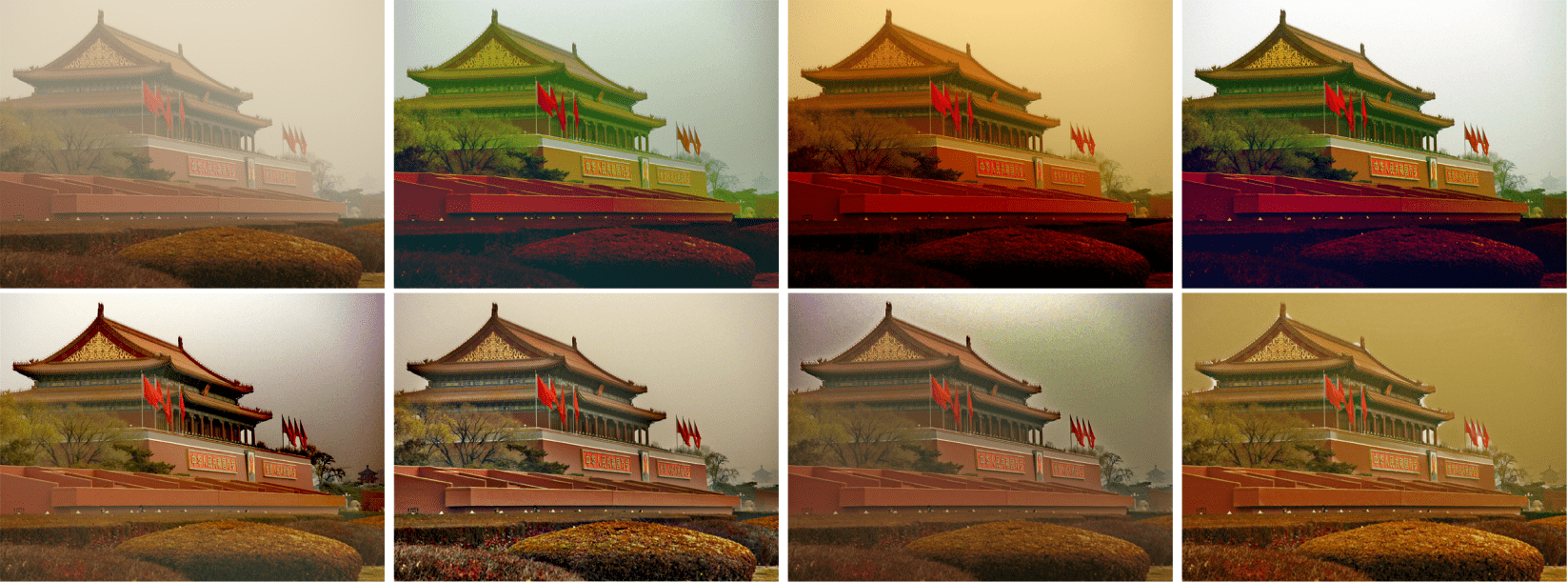}                
\caption{\textbf{Samples \#10 from our dataset with our methods compared to each other and to state-of-the-art methods.} First column represents the original image taken in bad weather. The next three columns represent from left to right, settings of $\lambda=1 - f_{init}$ using ace\cite{Gatta:2002}, clahe\cite{Zuiderveld:1994} and stress \cite{Kolas:2011} respectively. Bottom row with Berman \textit{et al.}\cite{Berman:2016}, Cho \textit{et al.}\cite{cho:2018}, Meng \textit{et al.}\cite{Meng:2013} and Tarel and Hauti{\`e}re \cite{Tarel:2009}}
\label{our_filterConf14}
\end{center}
\end{figure*}

\begin{figure*}[h!]
  \begin{center}
        \centering
                \includegraphics[width=.99\textwidth]{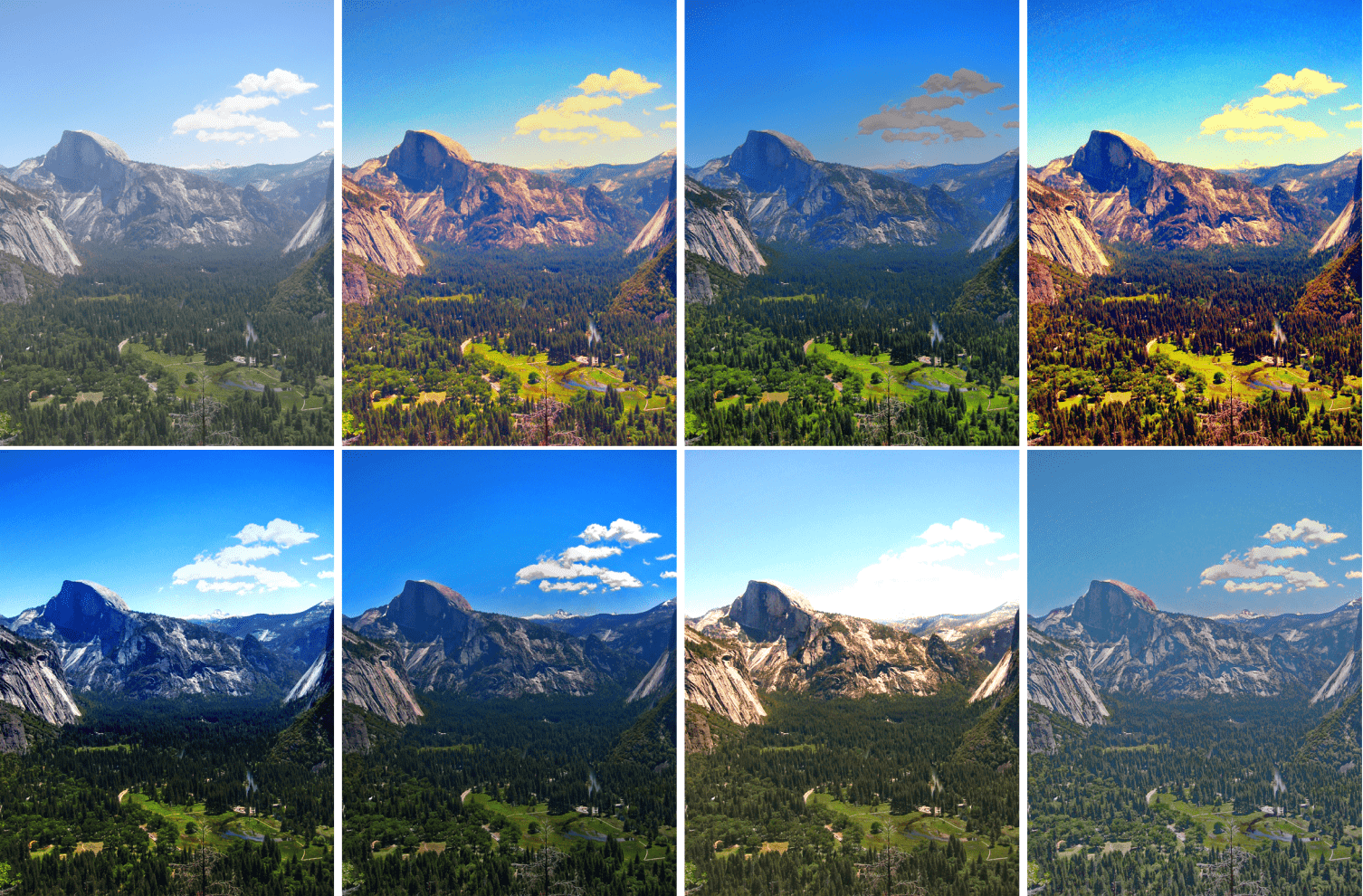}            
\caption{\textbf{Samples \#11 from our dataset with our methods compared to each other and to state-of-the-art methods.} First column represents the original image taken in bad weather. The next three columns represent from left to right, settings of $\lambda=0.35$ using ace\cite{Gatta:2002}, clahe\cite{Zuiderveld:1994} and stress \cite{Kolas:2011} respectively. Bottom row with  Choi \textit{et al.}\cite{Choi:2015}, He \textit{et al.}\cite{He:2011}, Meng \textit{et al.}\cite{Meng:2013} and Tarel and Hauti{\`e}re \cite{Tarel:2009}}
\label{our_filterConf1}
\end{center}
\end{figure*}

\begin{figure*}[h!]
  \begin{center}
        \centering
                \includegraphics[width=.99\textwidth]{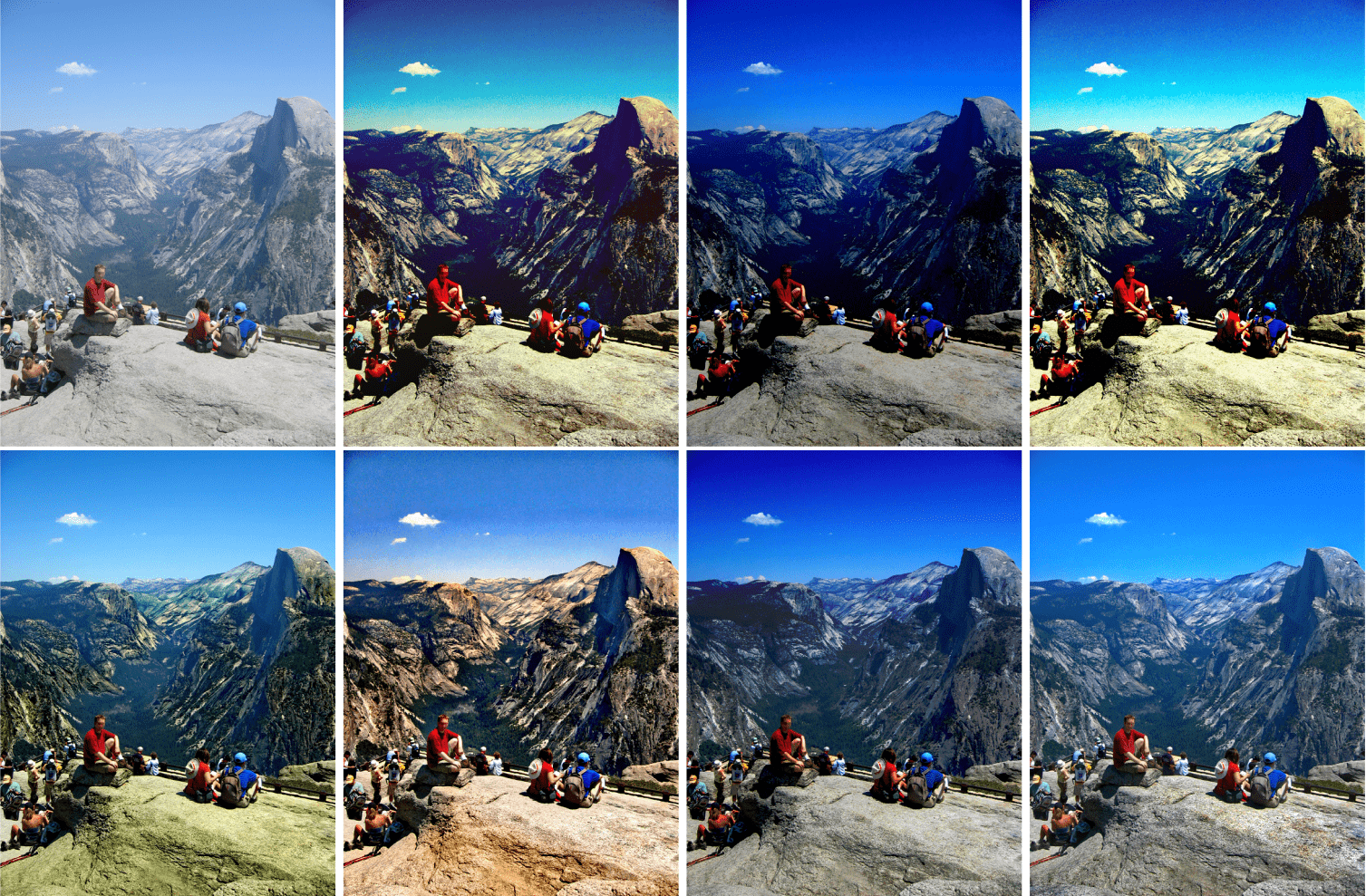}   
\caption{\textbf{Samples \#12 from our dataset with our methods compared to each other and to state-of-the-art methods.} First column represents the original image taken in bad weather. The next three columns represent from left to right, settings of $\lambda=1 - f_{init}$ using ace\cite{Gatta:2002}, clahe\cite{Zuiderveld:1994} and stress \cite{Kolas:2011} respectively. Bottom row with Berman \textit{et al.}\cite{Berman:2016}, Cho \textit{et al.}\cite{cho:2018}, Choi \textit{et al.}\cite{Choi:2015} and He \textit{et al.}\cite{He:2011}}
\label{our_filterConf1}
\end{center}
\end{figure*}

\section{Code}
All the code will be released soon.

\section{Miscellaneous}
This is a work in progress that we have not yet submitted in a review. However, it can already serve the community at this stage. Since we have not based most of our deduction on exhaustive experimentation, some of our conclusions need to be confirmed with a more formal approach.

\end{document}